\documentclass[twoside,11pt]{article}
\usepackage{jair, theapa, rawfonts}

\firstpageno{1}

\newcommand{\ignore}[1]{}

\newcommand{\blue}[1]{{\color{blue} #1}}

\newcommand{\orange}[1]{{\color{orange} #1}}

\usepackage{xcolor}
\usepackage{graphicx}
\usepackage{amsmath}
\usepackage{xspace}
\usepackage{comment}
\usepackage{url}
\newtheorem{definition}{Definition}
\DeclareMathOperator*{\argminB}{argmin}
\DeclareMathOperator*{\argmaxB}{argmax}
\usepackage{booktabs}
\usepackage{enumitem}
\setitemize{noitemsep,topsep=0pt,parsep=0pt,partopsep=0pt,leftmargin=*}

\newcommand{\methodt}{\textsc{Transmeter}\xspace}
\newcommand{\methodtS}{\textsc{Transmeter-S}\xspace}
\newcommand{\methodtR}{\textsc{Transmeter-R}\xspace}

\begin{document}

\title{Fast and Accurate Transferability Measurement for Heterogeneous Multivariate Data
}

\author{\name Seungcheol Park \email ant6si@snu.ac.kr \\
		\addr Seoul National University
       \AND
       \name Huiwen Xu \email xuhuiwen33@snu.ac.kr \\
       \addr Seoul National University
       \AND
       \name Taehun Kim \email taehun33.kim@samsung.com \\
       \addr Samsung Research, Samsung Electronics Co., Ltd.
       \AND
       \name Inhwan Hwang \email ihwan.hwang@samsung.com \\
       \addr Samsung Research, Samsung Electronics Co., Ltd.
       \AND
       \name KyungJun Kim \email kj424.kim@samsung.com \\
       \addr Samsung Research, Samsung Electronics Co., Ltd.
       \AND
       \name U Kang \email ukang@snu.ac.kr \\
       \addr Seoul National University}


\maketitle

\section*{Abstract}
Given a set of heterogeneous source datasets with their classifiers, how can we quickly find the most useful source dataset for a specific target task?
We address the problem of measuring transferability between source and target datasets, where the source and the target have different feature spaces and distributions.
We propose \methodt, a fast and accurate method to estimate the transferability of two heterogeneous multivariate datasets.
We address three challenges in measuring transferability between two heterogeneous multivariate datasets: reducing time, minimizing domain gap, and extracting meaningful homogeneous representations.
To overcome the above issues, we utilize a pre-trained source model, an adversarial network, and an encoder-decoder architecture.
Extensive experiments on heterogeneous multivariate datasets show that
\methodt gives the most accurate transferability measurement with up to 10.3$\times$ faster performance than its competitor. 
We also show that selecting the best source data with \methodt followed by a full transfer leads to the best transfer accuracy and the fastest running time.

\section{Introduction}
\label{sec:Intro}
Given a set of heterogeneous  source datasets with their classifiers, how can we quickly find the most useful source dataset for a specific target task?
In supervised learning, the amount of labeled data has a direct effect on the performance of a target task.
However, labeling a sufficient amount of data is costly,
and it is often impossible to get enough data when it comes to rare events or restricted data, e.g., mechanical faults or personal information.
For this reason, there have been growing interest in \emph{transfer learning} which aims to transfer data or model from a source to a target, to reduce the demand of the target data.
%
%
To exploit the advantage of the transfer learning, researchers enlarge the pool of source datasets
from using
\emph{homogeneous transfer learning}~\cite{Daume07,YaoD10,PanNSYC10,GongSSG12,OquabBLS14},
where both source and target domains have an identical feature space,
to using
\emph{heterogeneous transfer learning}~\cite{Shi2010,KulisSD11,WangM11,DuanXT12,ZhouTPT14,LiDXT14,YeSZH18}
where the two feature spaces are heterogeneous with different dimensions and distributions.

However, as the number of source candidates increases, it takes tremendous time to find the best source data for a given target task since we have to train the model using all source-target pairs.
Thus, we need to find a new algorithm that quickly estimates the performance gain of each source data, i.e. measuring transferability.
Measuring transferability between two heterogeneous datasets imposes the following challenges.
1) The algorithm needs to be fast enough to handle the extended size of the pool of source candidates.
Because of the heterogeneity between source and target datasets,
2) it is difficult to match the two domains, and 3) we may lose the key information during the domain transformation.

In this paper, we propose \methodt, a fast and accurate method that measures the transferability between source and target domains.
	The overall architecture of \methodt is depicted in Figure~\ref{fig:Transmeter}.
	\methodt consists of four modules: a label predictor for label prediction, a target encoder that maps the target domain to the homogeneous domain, a domain classifier that classifies whether the data come from the source or the target domain, and a target decoder that maps the homogeneous domain to the target domain.
	Note that \methodt performs an asymmetric feature transformation that transforms the dimension of the target data to that of the source data to reuse the pre-trained source model.
	To address the aforementioned challenges, we propose the following ideas:
	1) we initialize the weights of the label predictor using the pre-trained source model to reduce the training time,
	2) we form an adversarial architecture with the encoder to curtail the gap between the homogeneous representations of the source and the target domains, and
	3) to extract meaningful homogeneous representations, we
	introduce a target decoder that reconstructs original target data using corresponding representation in the homogeneous domain.
%
Extensive experiments on massive heterogeneous multivariate datasets show that \methodt
gives the most accurate and the fastest transferability measurement.

The main contributions of this paper are as follows.

\begin{figure}[t]
	\minipage{0.49\linewidth}
	\includegraphics[width=\linewidth]{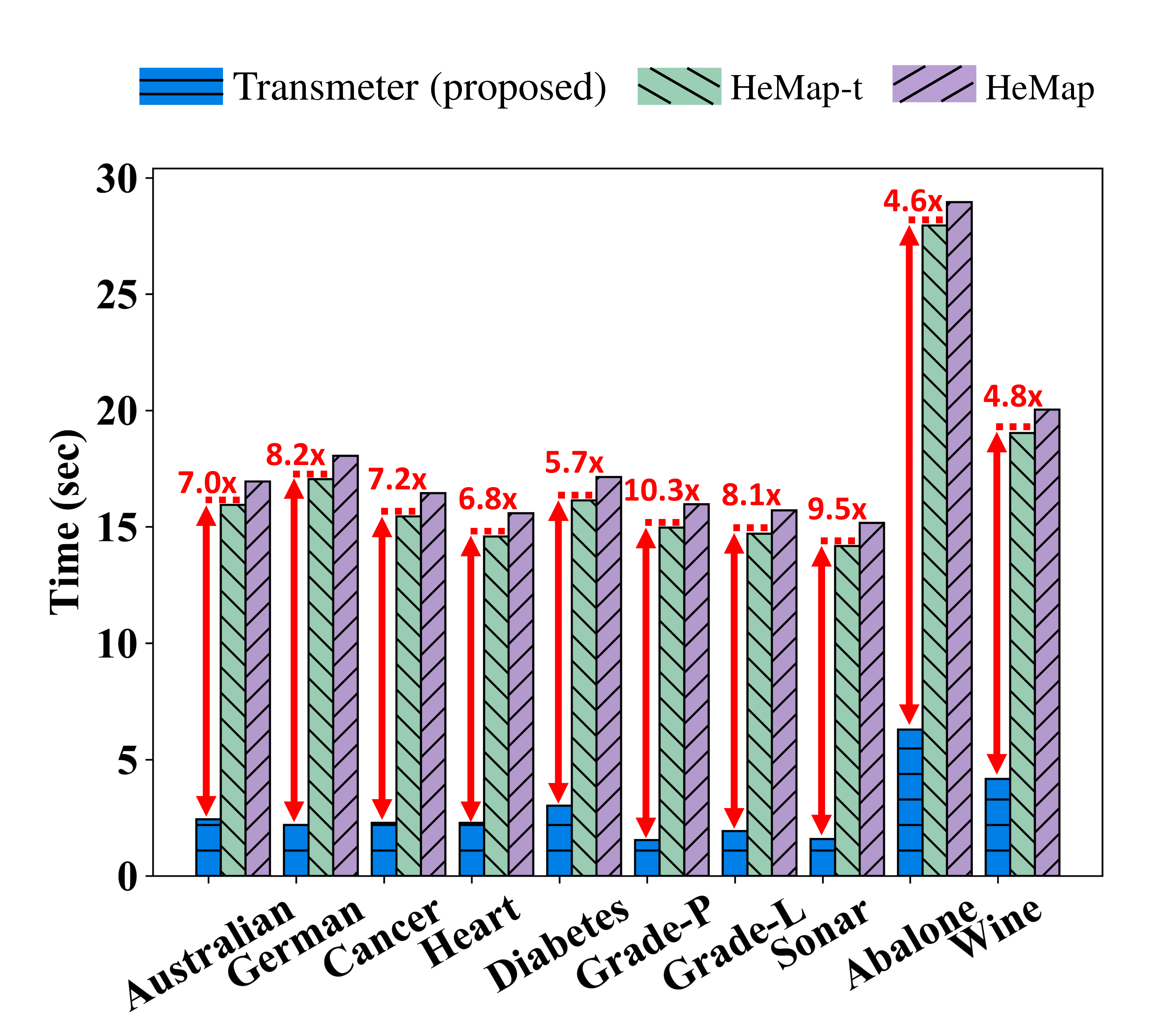}	\\	
	\caption{
		Comparison of training times of \methodt and competitors.
		\methodt shows up to 10.3$\times$ faster training time than that of the second best competitor. 
	}\label{fig:time}
	\endminipage\hfill
	\minipage{0.49\linewidth}
	\includegraphics[width=\linewidth]{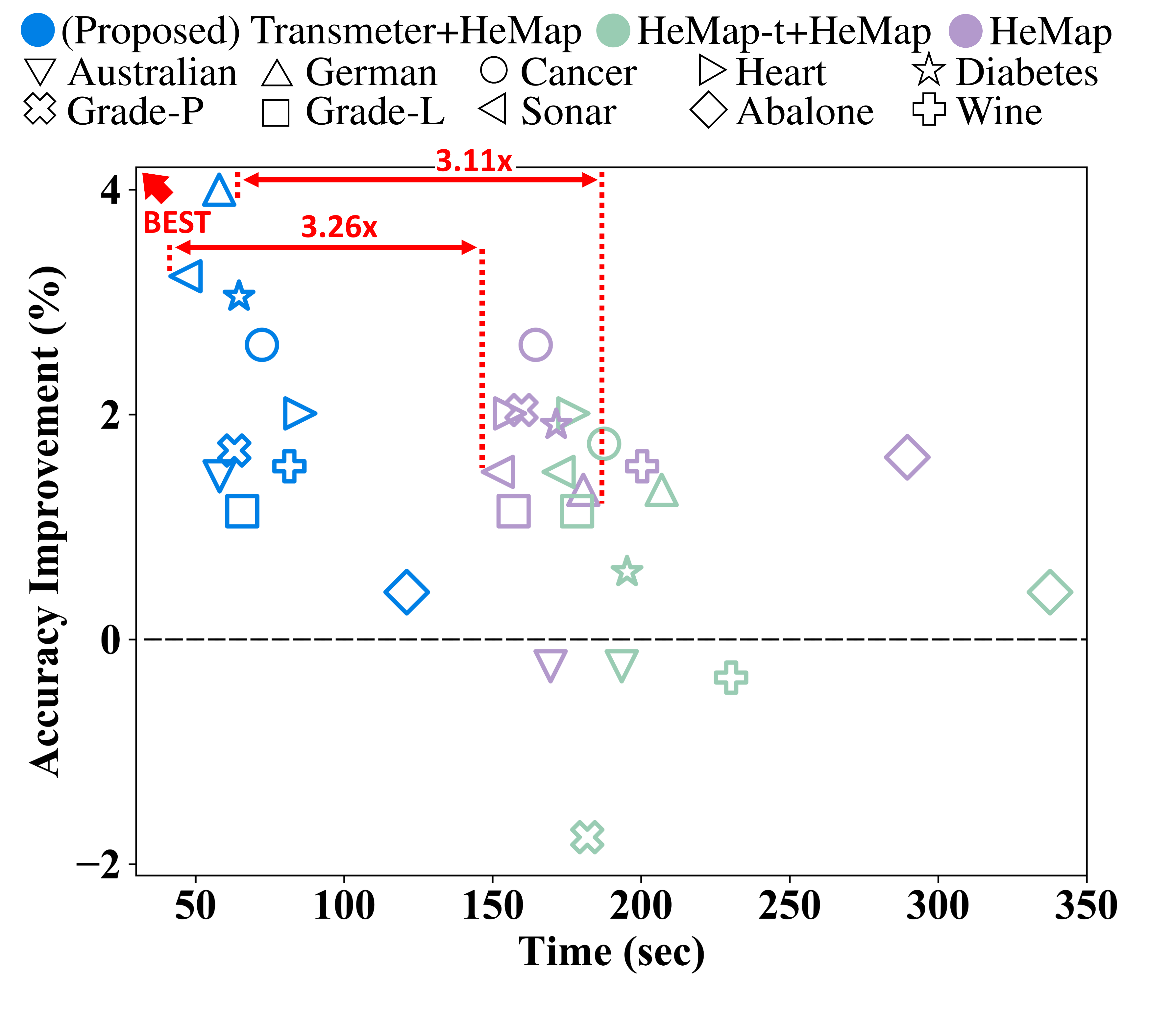}	\\
	\caption{
		Using \methodt to measure transferability, and fully transferring the selected data
		provides the best accuracy with the fastest running times.
	}\label{fig:scenario}
	\endminipage\hfill
\end{figure}

\begin{itemize}[label={-}]
	\item \textbf{Problem Definition.}
	We define the problem of measuring the transferability when working with heterogeneous multivariate datasets.
	While previous works focus on fully transferring knowledge, we focus on measuring \emph{transferability} between heterogeneous datasets.
%
%
   	\item \textbf{Method.} 	
	Our proposed method \methodt swiftly and accurately measures the transferability between two datasets by leveraging a pre-trained source model, domain classifier, and reconstruction loss to model a novel end-to-end solution.
%
    \item \textbf{Experiments.}
	We perform extensive experiments on heterogeneous multivariate datasets and show that
\methodt finds the best model 10.3$\times$ faster than competitors while minimizing accuracy reduction (see Figure~\ref{fig:time}).
We also show that selecting the best source data with \methodt followed by a full transfer leads to the best transfer accuracy and 3.26$\times$ faster running time than the second-best method (see Figure~\ref{fig:scenario}).
\end{itemize}

In the rest of this paper, we describe related works, proposed method, experiments, and conclusion in Sections~\ref{sec:Relateds}$\sim$\ref{sec:Conclusions}, respectively.

\ignore{
\begin{table}
\setlength{\tabcolsep}{3pt}
    \centering
 	\caption{Symbol description.}
 	\label{table:symbols}
	\begin{tabular}{clcl}
		\toprule
		\textbf{Symbol} 	& \textbf{Description} & \textbf{Symbol} 	& \textbf{Description}\\
		\midrule
		$N$				& Number of given source tasks
			& $y_s$, $y_t$ & Labels \\
		$D_s$			& A set of $N$ source datasets
			& $\hat{y}_s$, $\hat{y}_t$ & Predicted labels\\
		$H_s$			& A set of $N$ source classifiers
			& $d_s$, $d_t$ &	Domains \\
		$n_s$, $n_t$ 	& Number of the source and target instances
			& $\hat{d}_s$, $\hat{d}_t$& Predicted domains \\
		$x_s^{(i)}$		& Input features of the $i$-th source data
			& $\theta_e$ & Parameters of the target encoder\\
		$y_s^{(i)}$		& Labels of the $i$-th source data
			& $\theta_d$ & Parameters of the target decoder\\
		$h_s^{(i)}$		& Classifier of the $i$-th source data
			& $\theta_{lp}$ & Parameters of the label predictor\\
		$d_s^{(i)}$		& Input dimension of the $i$-th source data
			& $\theta_{dc}$ & Parameters of the domain classifier\\
		$x_{s,ds}$			& Input features of source data
			& $f_e$ & Target encoder\\
		$x_{t,dt}$			& Input features of target data
			& $f_d$ & Target decoder\\
		$x_{t,ds}$			& Encoded features of the target data in $d_s$
			& $f_{lp}$ & Label predictor\\
			$x'_{t,dt}$ & Reconstructed target data & $f_{dc}$ & Domain classifier\\

		\bottomrule
	\end{tabular}
\end{table}
} 

\section{Related Works}
\label{sec:Relateds}
We review previous works on heterogeneous transfer learning, relevance of source data, and domain adversarial learning.


\subsection{Heterogeneous Transfer Learning}

Heterogeneous transfer learning aims for transferring knowledge between heterogeneous domains. 
However, different feature spaces and distributions impose significant challenges such as severe negative transfer.
Recent studies to address these challenges are divided into two groups: symmetric and asymmetric feature-based transfer learning.
%
Symmetric approaches transform both source and target domains into a common latent space~\cite{DuanXT12,Shi2010,YanLNTWMW17} while asymmetric approaches transform only a source domain to a target domain~\cite{KulisSD11,ZhouTPT14,YeSZH18}.
\methodt belongs to the asymmetric feature-based transfer learning.
\methodt transforms the target domain into the dimension of the source domain to make homogeneous representations in the source domain.
This asymmetric feature transformation enables \methodt to reuse the pre-trained source classifier which accelerates its training.
%




\subsection{Relevance of Source Data}
The purpose of transfer learning is to improve the performance of a target learner by utilizing source data.
However, the target learner can be negatively impacted if the source domain is weakly related to the target domain, which is referred to as negative transfer~\cite{Rosenstein05,WeissK016}.
Hence, variant researches are performed to consider the relevance of source datasets in multi-source transfer learning~\cite{CaoL0J18,DuanXC12},
or to select relevant source data~\cite{Wang18} to avoid a negative transfer.
Cao \emph{et~al}. propose Supervised Local Weight (SLW) scheme to attenuate the effect of irrelevant source dataset~\cite{CaoL0J18},
and Wang \emph{et~al}. impose weights for each source data considering their relevance~\cite{Wang18}.
Our proposed method \methodt directly measures the transferability of each source dataset and enables us to find the most relevant one beyond merely avoiding the negative transfer.

\subsection{Domain Adaptation}

Domain adaptation aims to minimize the gap between source and target domains, which is crucial for the performance of the target learner after transferring source data. 
Ganin \emph{et~al}. propose Domain Adversarial Neural Network (DANN) that introduces the adversarial architecture of a domain classifier and a feature extractor~\cite{dann};
the domain classifier enforces the feature extractor to generate indistinguishable representations of source and target data.
The adversarial architecture of a domain classifier and a feature extractor is utilized in various recent researches in transfer learning~\cite{ChenCCTWS17,HoffmanWYD16,LongC0J18}.
\methodt adopts the adversarial architecture of a target encoder and a domain classifier to reduce the domain gap between homogeneous representations of source and target datasets.
Additionally, we introduce a target decoder that minimizes the reconstruction loss of the target data to make the target encoder extract meaningful homogeneous representations.


\section{Proposed Method}
\label{sec:Methods}
In this section, we formally define the problem and propose \methodt, a fast and accurate method for measuring transferability.

\subsection{Problem Definition}
\label{subsec:problem}

We define the problem of selecting the best source for a given target task. 
Given a set $D_s = \{ D_{s_1}, ..., D_{s_N}\}$ of $N$ source datasets with $N$ pre-trained models $F_{lp}=\{f_{lp_1}, ..., f_{lp_N}\}$ and a target dataset $D_t$,
our objective is to find the best source data and its related classifier that improve the target accuracy the most after transferring them to the target task.
We assume that all the pre-trained models have sufficient performance on their own domains. 
The $i$-th source dataset $D_{s_i}=\{(x^{j}_{s_i}, y^{j}_{s_i})\}^{n_{s_i}}_{j=1}$ has $n_{s_i}$ samples while the target dataset  $D_t=\{(x_t^j,y_t^j)\}^{n_t}_{j=1}$ has $n_t$ samples, where $x^{j}_{s_i}$ $\in$ $R^{d_{s_i}}$, $x_t^j$ $\in$ $R^{d_{t}}$, and $y_{s_i}^j$, $y_t^j$ $\in$ $\{ 0,1\}$.
We focus on
\textit{heterogeneous transfer learning} where $d_{s_i} \ne d_t$,
and
binary classification.

To precisely evaluate the performance gain, we define the transferability we aim to optimize.

\begin{definition}[Transferability]
	Given a pair of a source and a target data, the transferability between them is defined as the ratio of accuracy improvement by transferring the source data for the target task.
	\begin{equation}
	Transferability = \frac{acc_{T} - acc_{0}}{acc_{0}} \times 100\ (\%)
	\label{eq:transferability}
	\end{equation}
	where $acc_{T}$ and $acc_{0}$ represent the target accuracy with and without transferring the source data to the target task, respectively.
	\label{def:transferability}
\end{definition}

\begin{figure}[t]
                \centering
                \includegraphics[width=1.0\linewidth]{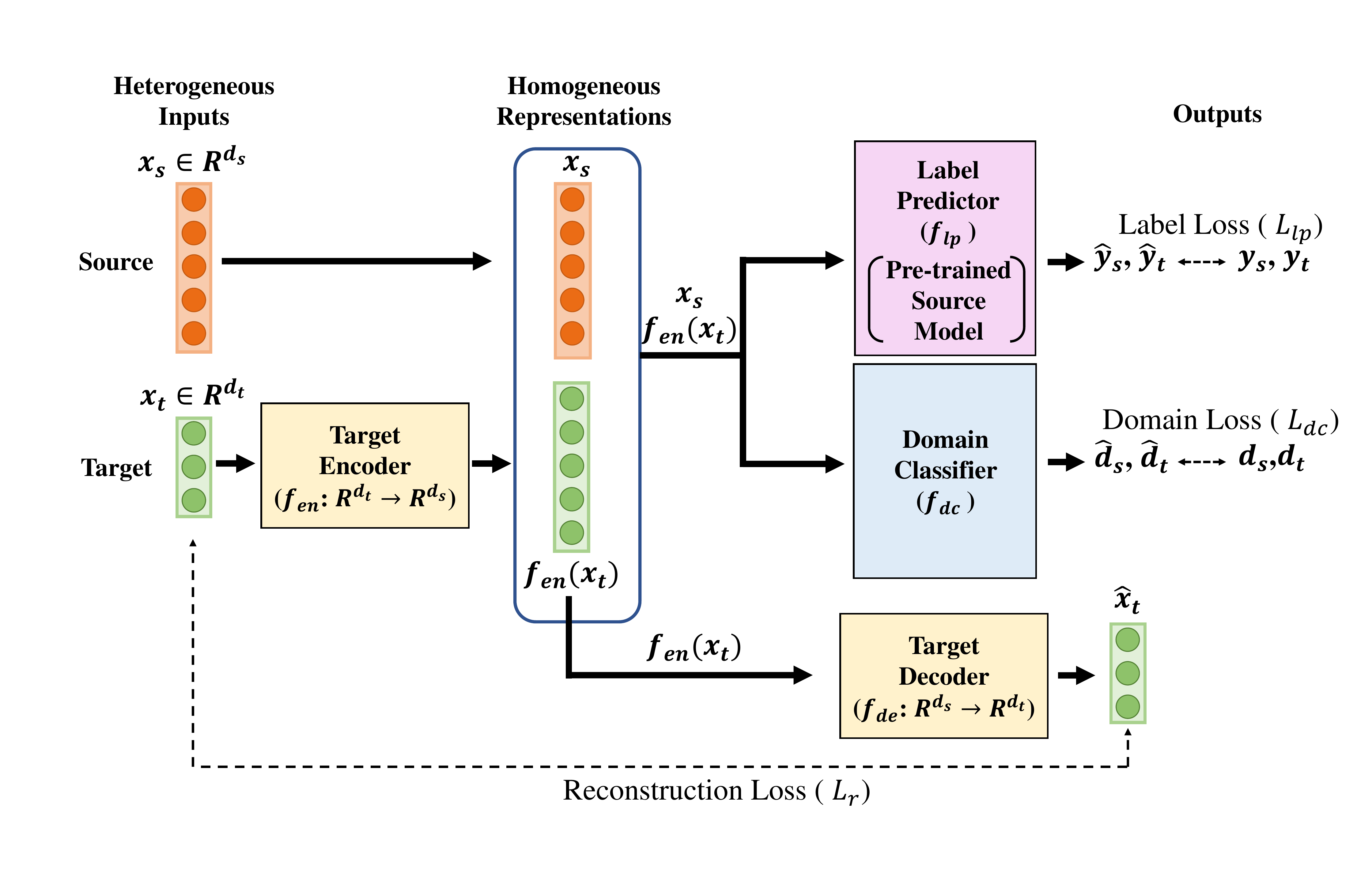}
                \caption{
\methodt consists of four modules: a target encoder that maps a target domain to a homogeneous domain, a target decoder that maps the homogeneous domain to the target domain, a label predictor for label prediction, and a domain classifier that classifies whether the data come from the source or the target domain. \methodt aims to transform the target data 
such that they are indistinguishable from the source data.
}
          		\label{fig:Transmeter}
\end{figure}

\subsection{Overview}
\label{subsec:overview}

Our proposed method \methodt evaluates the transferability between source datasets $\{ D_{s_i} \}$ and a target dataset $D_t$.
After evaluating the transferability of each source dataset, we consider the one with the highest transferability as the best source data.
As shown in Figure~\ref{fig:Transmeter},
\methodt consists of four modules: target encoder $f_{en}$, target decoder $f_{de}$, label predictor $f_{lp}$, and domain classifier $f_{dc}$.
We design \methodt to address the following challenges on measuring transferability between two heterogeneous multivariate datasets.
\begin{itemize}[label={-}]
	\item \textbf{Reducing time.}
	How can we quickly measure transferability to fully exploit the advantage of the transferability measurement algorithm?
	
	\item \textbf{Minimizing the domain gap.}
	It is more difficult to match two \emph{heterogeneous} datasets than \emph{homogeneous} datasets.
	How can we effectively reduce the domain gap between two heterogeneous datasets?

	\item \textbf{Extracting meaningful target features.}
How can we keep the crucial features of the target instances for accurate classification, while minimizing the domain gap?
\end{itemize}

We address the aforementioned challenges by the following ideas:

\begin{itemize}[label={-}]
	\item \textbf{Reusing a pre-trained source model.}
	We reuse the classifier pre-trained on the source data as the label predictor;
this gives a better initialization for the label predictor, and reduces the training time of \methodt.
	\item \textbf{Domain adversarial network.}
	We design a domain adversarial network that consists of a target encoder and a domain classifier to minimize the domain gap.
	The domain classifier aims to determine whether an instance comes from the source or the target domains.
The target encoder generates homogeneous representations to confuse the domain classifier.
This enables the target encoder to extract domain-invariant features which are fed into the label predictor to classify target instances.
	\item \textbf{Encoder-decoder architecture.}
	We introduce a target decoder, which forms an encoder-decoder architecture with the target encoder, that reconstructs the original target data from their homogeneous representations.
Combined with the domain adversarial network, this makes the homogenous representations of target instances keep their crucial features while being indistinguishable from those of source instances in terms of label predictor's point of view.
\end{itemize}

We design our objective function so that it includes the three main ideas above.
We explain the details of our objective function in the following section.

\subsection{Objective Function}

We define our learning objective as follows:

\begin{equation}
	\label{eq:loss}
	\begin{split}
		E(\theta_{en}, \theta_{de}, \theta_{lp}, \theta_{dc}) =
		\frac{1}{n_s+n_t} \left [ \sum ^{n_s}_{i=1} L^i_{lp}(x^i_s,y^i_s;\theta_{en},\theta_{lp}) + \sum ^{n_t}_{i=1} L^i_{lp}(x^i_t,y^i_t;\theta_{en},\theta_{lp}) \right ] \\
		- \alpha \cdot \frac{1}{n_s+n_t} \left [ \sum ^{n_s}_{i=1} L^i_{dc}(x^i_s,d^i_s;\theta_{en},\theta_{dc}) + \sum ^{n_t}_{i=1} L^i_{dc}(x^i_t,d^i_t;\theta_{en},\theta_{dc}) \right ] \\
		+ \beta \cdot \left[ \frac{1}{n_t} \sum ^{n_t}_{i=1} L^i_{r}(x^i_t;\theta_{en},\theta_{de}) \right]
	\end{split}
\end{equation}

$\theta_{en}$, $\theta_{de}$, $\theta_{lp}$, and $\theta_{dc}$ are trainable parameters of the target encoder, the target decoder, the label predictor, and the domain classifier, respectively.
The optimal parameters $\theta^*_{en}$, $\theta^*_{de}$, $\theta^*_{lp}$, and $\theta^*_{dc}$ satisfy

\begin{equation}
	\begin{split}
		(\theta^*_{en}, \theta^*_{de}, \theta^*_{lp}) &=\argminB_{\theta_{en}, \theta_{de}, \theta_{lp}} E(\theta_{en}, \theta_{de}, \theta_{lp}, \theta_{dc})\\
		\theta^*_{dc} &= \argmaxB_{\theta_{dc}} E(\theta_{en}, \theta_{de}, \theta_{lp}, \theta_{dc}) \\
	\end{split}
\end{equation}

The loss function $E$ in Equation~\ref{eq:loss} contains the loss terms
$L^i_{lp}$ for the label predictor (Section~\ref{subsec:lp}),
$L^i_{dc}$ for the domain classifier (Section~\ref{subsec:dc}), and
$L^i_{r}$ for the target feature reconstruction (Section~\ref{subsec:auto}).
$\alpha$ and $\beta$ are nonnegative hyperparameters that balance the loss terms.
$x^i_s$, $x^i_t$, $y^i_s$, and $y^i_t$ denote features and labels of the $i$-th source and target instances, respectively.
$d^i_s$ and $d^i_t$ are domain labels of $i$-th source instance and $i$-th target instance, respectively.
$n_s$ and $n_t$ are the numbers of the source and the target instances, respectively.

%

\subsubsection{Label Predictor}
\label{subsec:lp}


\setlength{\belowdisplayskip}{-0pt} \setlength{\belowdisplayshortskip}{-3pt}
\begin{equation}
	\label{eq:lp}
	L^i_{lp}(x^i,y^i;\theta_{en},\theta_{lp}) = -y^i \cdot log(\hat{y}^i)-(1-y^i) \cdot log(1-\hat{y}^i)
\end{equation}
\setlength{\belowdisplayskip}{-0pt} \setlength{\belowdisplayshortskip}{-2pt}
\begin{equation}
	\label{eq:y}
	\hat{y}^i = f_{lp}(Ho(x^i;\theta_{en});\theta_{lp})
\end{equation}
\setlength{\belowdisplayskip}{-0pt} \setlength{\belowdisplayshortskip}{8pt}
\begin{equation}
	\label{eq:ho}
	Ho(x^i;\theta_{en})= \left\{
		\begin{array}{ll}
			x^i 						&	if\ source\ data,\\
			f_{en}(x^i;\theta_{en})		&	otherwise\\
		\end{array}
	\right.
\end{equation}
$L^i_{lp}$ is the label predictor loss of the $i$-th instance.
$\hat{y}^i$ (Equation~\ref{eq:y}) denotes the predicted label of the $i$-th instance.
$Ho$ in Equation~\ref{eq:ho} denotes the homogeneous representation shown in Figure~\ref{fig:Transmeter}.
$f_{lp}$ is the label predictor with parameter $\theta_{lp}$, and $f_{en}$ is the target encoder with parameter $\theta_{en}$.

%

\subsubsection{Domain Classifier}
\label{subsec:dc}

The domain discrimination loss in Equation~\ref{eq:dc} is designed to
improve the accuracy of the target label prediction by
making the source and the target features indistinguishable.

\setlength{\belowdisplayskip}{-0pt} \setlength{\belowdisplayshortskip}{-3pt}
\begin{equation}
	\label{eq:dc}
	L^i_{dc}(x^i,d^i;\theta_{en},\theta_{dc}) = -d^i \cdot log(\hat{d}^i) -(1-d^i) \cdot log(1-\hat{d}^i)	
\end{equation}
\setlength{\belowdisplayskip}{-0pt} \setlength{\belowdisplayshortskip}{8pt}
\begin{equation}
	\label{eq:d}
	\hat{d}^i = f_{dc}(Ho(x^i;\theta_{en});\theta_{dc})	
\end{equation}

$L^i_{dc}$ is the domain classifier loss of the $i$-th instance.
$f_{dc}$ is the domain classifier with parameter $\theta_{dc}$, and $\hat{d}^i$ (Equation~\ref{eq:d}) denotes the predicted domain class of the $i$-th instance.


\subsubsection{Feature Reconstruction}
\label{subsec:auto}

The feature reconstruction loss in Equation~\ref{eq:r} is designed to extract useful features which are able to reconstruct the original target data.
This can be thought of as an autoencoder
where the encoder maps the input features into a constrained code, and the decoder recovers the code to the input features.
The model is trained to minimize the reconstruction loss for each target data point.

	
%

\setlength{\belowdisplayskip}{-0pt} \setlength{\belowdisplayshortskip}{-3pt}
\begin{equation}	
	\label{eq:r}
	L^i_{r}(x^i;\theta_{en},\theta_{de}) = || \hat{x}^i - x^i || ^2	
\end{equation}
\setlength{\belowdisplayskip}{-0pt} \setlength{\belowdisplayshortskip}{8pt}
\begin{equation}
	\label{eq:x}	
	\hat{x}^i = f_{de}(Ho(x^i;\theta_{en});\theta_{de})	
\end{equation}
$L^i_{r}$ is the feature reconstruction loss of the $i$-th instance.
$f_{de}$ is the domain classifier with parameter $\theta_{de}$, and $\hat{x}^i$ in Equation~\ref{eq:x} denotes the reconstructed feature of the $i$-th instance.


\subsection{\methodt}
Given a source dataset with its pre-trained model,
and a target dataset,
\methodt learns the model parameters $\theta_{en}$, $\theta_{de}$, $\theta_{lp}$, and $\theta_{dc}$ to measure the transferability.
Note that the parameters of the pre-trained source model are used for initializing the weight $\theta_{lp}$ of the label predictor.
In the training process, both the source and the target data are used to update the parameters. 

	We also consider the correspondence between labels of the source and the target data to maximize the performance of \methodt.
	We observe that 0-labeled source data often correspond to 1-labeled target data rather than 0-labeled target data.
	Thus, we design a hyperparameter for flipping target labels, and find the better alignment of labels by training twice for each source and target pair of datasets: flipping or not flipping target labels.

\section{Experiments}
\label{sec:Experiments}

We conduct experiments to answer the following questions on the performance of \methodt.
\begin{itemize}[label={-}]
	\item \textbf{Q1.$\;$Accuracy of transferability (Section \ref{sec:exp1}).}
	Does \methodt accurately predict the best source data?
	\item \textbf{Q2.$\;$Training time for transferability (Section \ref{sec:exp2}).}
	How much training time is saved by using \methodt for transferability measurement?
	\item \textbf{Q3.$\;$Exploiting transferability for full transfer (Section \ref{sec:exp3}).}
	How useful is \methodt for transfer learning with multiple source candidates?
	\item \textbf{Q4.$\;$Ablation study (Section \ref{sec:exp4}).}
	Does using a pre-trained source model and reconstruction loss improve the performance of \methodt?
\end{itemize}

\subsection{Experimental Settings}
\label{sec:exp_setting}

We describe experimental settings including datasets and competitors.
All codes are written in python 3.6 using PyTorch, and
all experiments are done in a workstation with Geforce GTX 1080 Ti. 

\paragraph{\textbf{Datasets.}}
We use ten multivariate datasets for binary classification summarized in Table~\ref{table:datasets}:
Australian\footnote{https://archive.ics.uci.edu/ml/datasets/Statlog+(Australian+Credit+Approval)}, German\footnote{https://archive.ics.uci.edu/ml/datasets/statlog+(german+credit+data)},
Cancer\footnote{http://archive.ics.uci.edu/ml/datasets/breast+cancer+wisconsin+(diagnostic)},
Heart\footnote{http://archive.ics.uci.edu/ml/datasets/Heart+Disease},
Diabetes\footnote{https://www.kaggle.com/uciml/pima-indians-diabetes-database},
Grade-P\footnote{https://www.kaggle.com/dipam7/student-grade-prediction},
Grade-L\footnote{https://www.kaggle.com/aljarah/xAPI-Edu-Data},
Sonar\footnote{http://archive.ics.uci.edu/ml/datasets/connectionist+bench+(sonar,+mines+vs.+rocks)},
Abalone\footnote{https://archive.ics.uci.edu/ml/datasets/Abalone},
and Wine\footnote{https://www.kaggle.com/uciml/red-wine-quality-cortez-et-al-2009}.
	Note that all the datasets are heterogeneous; they have different categories and lie on different feature spaces.
We split each data into training and test sets by 7:3 ratio, 
and preprocess them using z-normalization. 
%
\begin{table*}[h]
	\centering
	\label{table:datasets}
	\setlength{\tabcolsep}{1pt}
	\begin{tabular}{lllcc  }
		\toprule
		\textbf{Data Name} & \textbf{Abbr.}& \textbf{Category} & \textbf{Features} & \textbf{Instances} \\
		\midrule
		Australian Credit Approval\footnotemark[1]	& Australian	     & Financial & 14       & 690 	    \\
		German Credit data\footnotemark[2]	& German	     & Financial & 24       & 1000 	    \\
		Breast Cancer Wisconsin (Diagnostic)\footnotemark[3] & Cancer & Health    & 30   	& 569      	\\
		Heart Disease Data Set\footnotemark[4] & Heart & Health    & 13   	& 303      	\\
		Pima Indians Diabetes Database\footnotemark[5] & Diabetes & Health    & 8   	& 768      	\\
		Student Grade Prediction (Portugese)\footnotemark[6] & Grade-P			 & Education & 32 		& 395 		\\
		Student' Academic Performance Dataset\footnotemark[7] & Grade-L & Education   & 16   	& 480      	\\
		Connectionist Bench (Sonar, Mines vs. Rocks)\footnotemark[8] & Sonar & Science    & 60   	& 208     	\\
		Abalone Data set\footnotemark[9] & Abalone & Science    & 8   	& 4177    	\\
		Red Wine Quality\footnotemark[10] & Wine & Science    & 11   	& 1599   	\\
		
		\bottomrule
	\end{tabular}
	\caption{Summary of the datasets.}
\end{table*}

\paragraph{\textbf{Competitors.}}
We use HeMap~\cite{Shi2010} and its variant as competitors.
HeMap~\cite{Shi2010} is the most recent heterogeneous transfer learning method for multivariate data.
It samples source data points near target data points and determines that it is too risky to transfer when the ratio of the selected source data is lower than a threshold.
We exploit this ratio of the selected source data as a transferability between two datasets and name this method as HeMap-t.
We utilize HeMap for fully transferring source data after finding the best source data and leverage HeMap-t for a competitor of \methodt as a transferability measurement algorithm.

\paragraph{\textbf{Hyperparameters.}}
We use four or five hidden layers for pre-trained source models, three or four hidden layers for the target encoder, and a single layer for the domain classifier.
The architecture of the target decoder is symmetric to that of the target encoder, and the label predictor has the identical structure with the pre-trained source model.
We use MLPs for all modules since we focus on multivariate data and
	apply He-initialization~\cite{HeZRS15}, ReLU activation, and batch normalization as a default setting for each layer.
We select the best values of $\alpha$ and $\beta$ among $\alpha \in $ $\{$0.003, 0.01, 0.1, 0.3, 1.0, 10.0$\}$ and $\beta \in \{0.1, 0.5\}$ to balance the loss terms for each dataset;
we use a larger set for $\alpha$ than $\beta$ since \methodt is sensitive more to the domain classifier.
We select the best random seed among $\{1,2,3,4,5\}$ to initialize random functions.
We perform 3-fold cross-validation to select the best set of hyperparameters and use Adam optimizer for training all methods.

\subsection{Accuracy of Transferability}
\label{sec:exp1}

We evaluate the transferability of \methodt and the competing method HeMap-t.
We measure the transferability of each source data and select the top $k$ sources ($k=1..4$); more than $k$  sources may be selected when the scores are the same.
Then, we check if the pool of selected source datasets includes the best source for HeMap, the full transfer method.
The results are summarized in Table~\ref{table:rank}
where the 2nd and the 3rd rows denote the number of correctly predicted best source models for 10 target datasets by utilizing \methodt and HeMap-t, respectively. 
For each target dataset, we use the other datasets as sources. 
Note that \methodt outperforms HeMap-t in all cases,
correctly finding the best source data in 80\% or 90\% of cases in Top-2, 3, 4 settings.

Additionally, we count the number of cases that target labels are flipped when we train the final model to verify the effect of label matching.
As a result, almost half of total cases (95 cases among 180 cases) show better performance when they are trained with flipped labels.
Thus, it is worthwhile to train each pair of source and target data twice with flipped and unflipped target labels considering the relationship between source and target labels.

\begin{table*}[h]
	\centering		
	\label{table:rank}
	\begin{tabular}{lcccc}
		\toprule
		\textbf{Methods}  & \textbf{Top-1} & \textbf{Top-2} & \textbf{Top-3} & \textbf{Top-4} \\
		\midrule
		\methodt & 5/10 & 8/10 & 8/10 & 9/10\\
		HeMap-t  & 4/10 & 5/10 & 7/10 & 9/10\\
		\bottomrule
	\end{tabular}
	\caption{
		Ratio of target tasks where their best source data are included in the top-$k$ sources found by transferability measurement methods. \methodt outperforms HeMap-t in all cases.
	}
\end{table*}

\subsection{Training Time}
\label{sec:exp2}

Figure~\ref{fig:time} shows the comparison of training times of \methodt and competitors.
We use an early stopping technique for training each model; 
we stop training when the number of consecutive epochs with increasing validation loss 
surpasses a threshold. 
%
%
	Note that \methodt shows up to 10.3 $\times$ faster training time than the second-best competitor HeMap-t. 
The short training time of \methodt enables measuring the transferability of all source candidates before performing a full transfer.

\subsection{Exploiting Transferability for Full Transfer}
\label{sec:exp3}

We evaluate how to exploit the transferability for quickly and accurately improving a target task via full transfer.
There are two ways to quickly increase the accuracy of the target task. 
First, we can use a transferability measurement method to find the best source, and use a full transfer algorithm to transfer the best source data to the target task.
Second, we can use a full transfer method to fully transfer all source data, and find the best one among them.
Thus, we compare the following three settings.
Note that we use 1) HeMap as a full-transfer method,
and 2) \methodt and HeMap-t as transferability measurement methods.

	\begin{itemize}[label={-}]
		\item \textbf{(Proposed) \methodt + HeMap.} We measure the transferability of each source data using \methodt, and select the top two source data; we include all the source data when there are multiple data with the same ranking.
We then transfer the selected source data to the target task using HeMap.
Although \methodt is designed as a transferability measurement method,
\methodt can be used for full transfer as well; thus
we choose the best model among the ones trained by HeMap and \methodt. 
		\item \textbf{HeMap-t + HeMap.} We measure the transferability of each source data using HeMap-t, and select the top two source data; we include all the source data when there are multiple data with the same ranking.
We then transfer the selected source data to the target task using HeMap.
Note that HeMap-t is not used for full transfer since it is designed only for transferability measurement.
		\item \textbf{HeMap.} We fully transfer all source data one by one using HeMap, and find the best model among them.
	\end{itemize}

The results of the experiments are shown in Figure~\ref{fig:scenario} where each color represents a method, and each symbol represents a target dataset.
The vertical axis represents the improvement of the target accuracy compared to the baseline trained only with each target data.
%
%
Note that using \methodt as a transferability measurement method
provides the best accuracy while minimizing running times.
\methodt always gives better or equal accuracy in identifying the best source data compared to HeMap-t, with faster running times.
\methodt combined with HeMap also provides better accuracy and smaller running times compared to an exhaustive search using only HeMap; \methodt+HeMap gives better or equal accuracy in 8 out of 10 cases, while running up to 3.26$\times$ faster than HeMap.

\subsection{Ablation Study}
\label{sec:exp4}
We conduct an ablation study to verify the contribution of using a pre-trained source model and the reconstruction loss (Equation~\ref{eq:r}).
We compare the training times and top-2 accuracy of \methodt with its following two variants.
\methodtS is a variant of \methodt that does not initialize weights of label predictor using a pre-trained source model; instead, \methodtS uses the default initialization method~\cite{HeZRS15} for linear layers in PyTorch.
\methodtR is a variant of \methodt trained without the reconstruction loss.
The result of ablation study in Table~\ref{table:ablation} explicitly shows that using the pre-trained source model for weight initialization accelerates the training of the \methodt with better accuracy,
and introducing the reconstruction loss greatly increases the accuracy of measuring transferability.

\begin{table*}[h]
	\centering
	\label{table:ablation}
	\begin{tabular}{lcc}
		\toprule		
		\textbf{Methods}                                                         & \textbf{Top-2 Acc.}& \textbf{Training Time (sec)}\\
		\midrule
		\methodt& 8/10 & 2.78  \\
		\methodt-S  & 7/10 & 7.87  \\
		\methodt-R & 2/10 & 2.25  \\ \bottomrule
	\end{tabular}
	\caption{
		\methodt gives the best trade-off of accuracy and training time, compared to its variants
		without the pre-trained source model (-S) and without the reconstruction loss (-R).
	}
\end{table*} 

\section{Conclusion}
\label{sec:Conclusions}

In this paper, we propose \methodt, a novel algorithm that measures the transferability between two heterogeneous multivariate datasets. 
We exploit
a pre-trained source model as a label predictor to reduce the training time,
an adversarial network to reduce the gap between the source and the target domains, and
an encoder-decoder architecture to extract meaningful homogeneous representations.
Extensive experiments on heterogeneous multivariate data demonstrate that
1) \methodt provides the most accurate transferability measurement with up to 10.3$\times$ faster training time than its competitor,
and
2) using \methodt to measure the transferability and fully transferring the selected data provides the best accuracy with the fastest running times.
Future works encompass extending the method for simultaneously transferring multiple heterogeneous source data. 

\acks{
	This work is supported by SAMSUNG Research,	
	Samsung Electronics Co.,Ltd.
}



\vskip 0.2in
\bibliography{ref}
\bibliographystyle{theapa}

\end{document}